\documentclass[10pt,twocolumn,letterpaper]{article}

\usepackage[pagenumbers]{cvpr} 

\usepackage{graphicx}
\usepackage{amsmath}
\usepackage{amssymb}
\usepackage{booktabs}

\usepackage[pagebackref,breaklinks,colorlinks]{hyperref}

\usepackage[capitalize]{cleveref}
\crefname{section}{Sec.}{Secs.}
\Crefname{section}{Section}{Sections}
\Crefname{table}{Table}{Tables}
\crefname{table}{Tab.}{Tabs.}

\def\cvprPaperID{*****} 
\def\confName{CVPR}
\def\confYear{2022}

\newcommand{\norm}[1]{\left\lVert#1\right\rVert}
\newcommand{\argmin}{\mathop{\mathrm{arg min}}\limits}

\usepackage{multirow}
\usepackage{mathtools,cuted}
\usepackage{subcaption}

\begin{document}

\title{OSOP: A Multi-Stage One Shot Object Pose Estimation Framework}

\author{%
	Ivan Shugurov$^{1, 3}$, Fu Li$^{1, 2}$, Benjamin Busam$^{1}$, Slobodan Ilic$^{1, 3}$ \\
	\\ $^{1}$ TU Munich \quad $^{2}$ NUDT \quad $^{3}$ Siemens AG \\
	\texttt{\{ivan.shugurov, fu.li, b.busam, slobodan.ilic\}@tum.de}
}
\maketitle

\begin{abstract}
    We present a novel one-shot method for object detection and 6 DoF pose estimation, that does not require training on target objects. At test time, it takes as input a target image and a textured 3D query model. The core idea is to represent a 3D model with a number of 2D templates rendered from different viewpoints. This enables CNN-based direct dense feature extraction and matching. The object is first localized in 2D, then its approximate viewpoint is estimated, followed by dense 2D-3D correspondence prediction.  The final pose is computed with PnP.  We evaluate the method on LineMOD, Occlusion, Homebrewed, YCB-V and TLESS datasets and report very competitive performance in comparison to the state-of-the-art methods trained on synthetic data, even though our method is not trained on the object models used for testing.
\end{abstract}

\section{Introduction}
The rapid development of high quality 6 DoF pose estimation methods is underway. According to the BOP challenge~\cite{bopchallenge}, which combines publicly available 6 DoF pose estimation datasets and offers standardized evaluation and comparison procedures, the field is dominated by deep learning methods~\cite{zakharov2019dpod,labbe2020,li2019cdpn,park2019pix2pose,song2020hybridpose,hodan2020epos,peng2019pvnet,kaskman2020,shugurov2021,sundermeyer2018implicit,sundermeyer2020multi,kehl2017ssd,jafari2018ipose,brachmann2014learning,shugurov2021dpodv2,labbe2020,li2022polarmesh,li2022ws,li2022nerf}.  The methods' performance is, however, limited by the availability of labeled training data. Accurate 6 DoF pose annotation of real data is a complicated and time-consuming process~\cite{Kaskman_2019_ICCV_Workshops}, that must be repeated manually for each new object. This severely limits the practical applicability of 6 DoF pose estimation methods. Additionally, labels can exhibit imperfection~\cite{busam2020like}. As synthetic data preparation tools improved, more methods shifted to training on synthetically rendered images. This greatly simplifies the preparation of data for new objects. These time-consuming and computationally-intensive data rendering and model training steps, on the other hand, must be repeated for each new target object of interest.

\begin{figure}
\centering
\begin{subfigure}{.45\linewidth}
  \centering
  \includegraphics[width=0.99\linewidth]{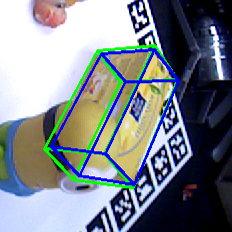}
  \caption{\label{fig:vis_11}}
  
  \includegraphics[width=0.99\linewidth]{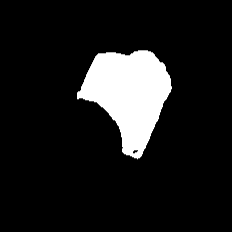}

  \caption{\label{fig:vis_21}}
  
\end{subfigure}%
\begin{subfigure}{.45\linewidth}
  \centering

  \includegraphics[width=0.99\linewidth]{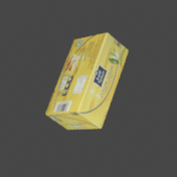}
  
  \caption{\label{fig:vis_12}}
  
  \includegraphics[width=0.99\linewidth]{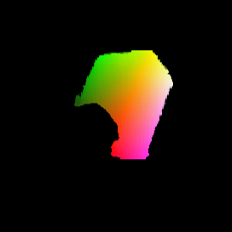}
  
  \caption{\label{fig:vis_22}}
  
\end{subfigure}%
\vspace{-1em}
\caption{Qualitative evaluation of the proposed method on an object from the Homebrewed dataset~\cite{Kaskman_2019_ICCV_Workshops}. a) an input image, cropped only for visualization purposes, with a comparison of the ground truth (green cuboid) and estimated (blue cuboid) poses; b) predicted one-shot segmentation ; c) a matched template; and d) predicted correspondences as color-coded NOCS correspondences.\label{fig:visual_results2}}
\vspace{-2em}
\end{figure}

While one-shot object detection, i.e., detection of novel objects not seen during training, appears to yield promising results for conventional 2D detection, its extensions to pose estimation were very limited. 
There are very few related works that attempt to generalize to new objects. They primarily focus on objects from the same category~\cite{wang2019normalized,manhardt2020cps,chen2020category}, objects with very similar geometry~\cite{pitteri2019cornet,pitteri20203d}, rely on partially training on target objects~\cite{sundermeyer2018implicit}, or limit the task to viewpoint estimation~\cite{xiao2019pose}.  We extend the one-shot object detection ideas to estimate the full 6 DoF pose. Our method is trained only once and then automatically generalizes to new objects without training on them, obviating the need for synthetic or real data preparation and training for new objects. 

\begin{figure*}[!t]
    \centering
    \includegraphics[width=0.85\linewidth]{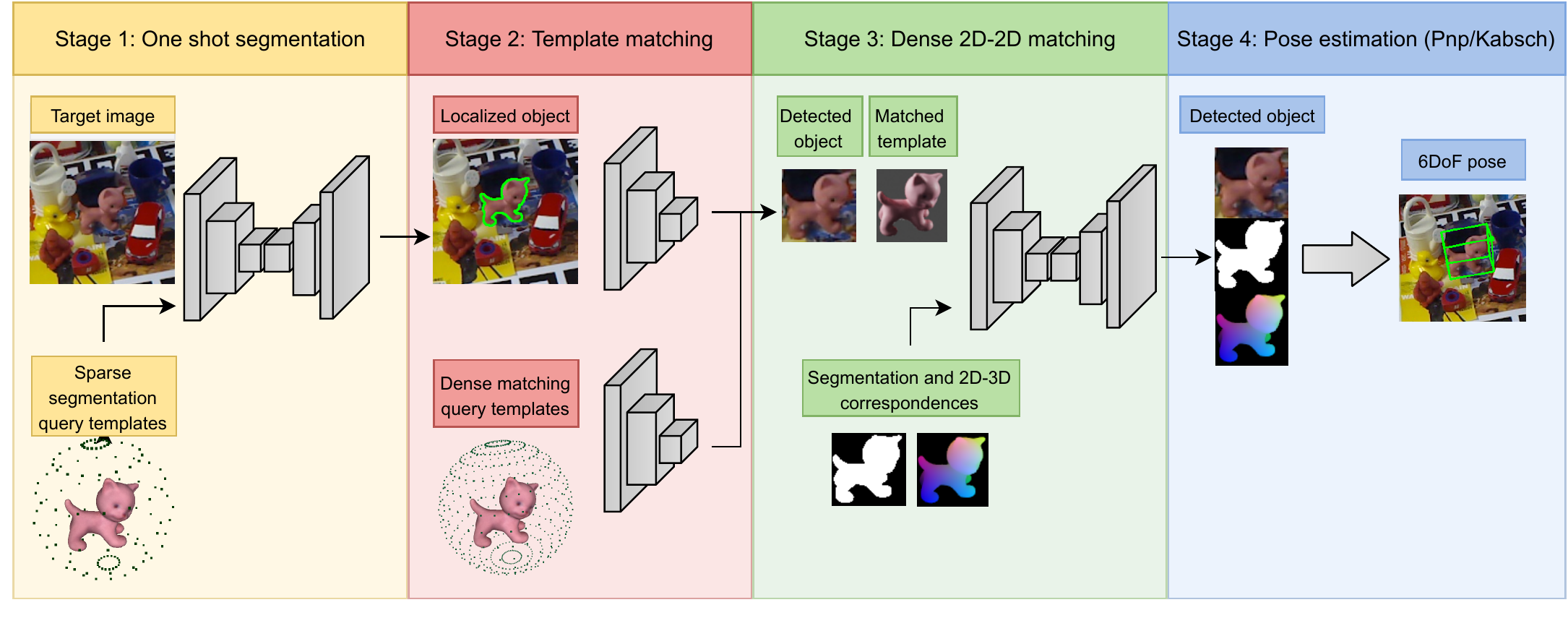}
    \vspace{-1em}
    \caption{\textbf{Pipeline of the proposed detector.} 1) One-shot object localization conditioned on the 3D model. 2) Initial viewpoint estimation by template matching. 3) Dense 2D-2D matching between the image patch and the matched template. 4) 6 DoF pose estimation with PnP+RANSAC or Kabsch+RANSAC. The proposed pose estimation pipeline generalizes well to new target objects not seen duing training.}
    \label{fig:pipeline}
    \vspace{-1.5em}
\end{figure*}

The 4-stage pipeline of the proposed approach is visualized in Figure~\ref{fig:pipeline}. The input to the method is a test image and a textured 3D model of the target object of interest. Inspired by OS2D~\cite{osokin2020os2d}, the method relies on dense sliding window-based feature correlation between the object, represented with a sparse set of 2D templates obtained by rendering the object from various viewpoints, and the input test image.  In the first stage, one-shot object segmentation is performed. The detected object is matched to a database of object renderings in the second stage to perform initial viewpoint estimation. In the third stage, a network estimates dense 2D-2D correspondences between pixels of the input image patch and the matched template, whose pose is known. This provides us with 2D-3D correspondences between the pixels of the input image and the 3D model. This enables 6 DoF pose estimation using PnP\cite{lepetit2009epnp} or Kabsch\cite{besl1992method} with RANSAC~\cite{fischler1981random} in the last stage.

The evaluation of our approach on five datasets (LineMOD, Occlusion, HomebrewDB, YCB-V and TLESS) proves that it fully generalizes to new objects and scenes not seen during training. Our key contributions include: 1) The first RGB-based one-shot pose estimation pipeline that truly scales to new objects without training on them. This results in %
a considerable time reduction required for synthetic data generation and retraining.
2) A novel architecture and a novel attention mechanism for one-shot semantic segmentation; 3) An architecture for dense 2D-2D matching that enables 2D-3D correspondence transfer from a template with known 6 DoF pose to a target image with unknown pose.
\section{Related Work}

{\textbf{State of the Art DL Methods.}} The main trend is to predict the locations of 2D keypoints for which their corresponding points on the 3D model are known. In particular, IPose~\cite{jafari2018ipose}, YOLO6D~\cite{tekin2018real}, PVNet~\cite{peng2019pvnet}, HybridPose~\cite{song2020hybridpose} and~\cite{kaskman2020} predict a sparse set of the pre-defined keypoints. DPOD~\cite{zakharov2019dpod,shugurov2021dpodv2,shugurov2021multi}, CDPN~\cite{li2019cdpn}, Pix2Pose~\cite{park2019pix2pose}, and EPOS~\cite{hodan2020epos} rely on dense pixel-wise 2D-3D correspondence estimation. Pose is then estimated using the PnP~\cite{lepetit2009epnp} algorithm. Prediction of keypoints allows for more robust pose estimation and explicitly uses handling of occlusions. A notable exception is CosyPose~\cite{labbe2020}, where excellent pose estimation results are obtained with direct pose regression followed by a multi-view refinement.\\

{\textbf{Generalization to Novel Objects.}}
Point Pair Features (PPF)~\cite{drost2010model} is arguably the only commercially available~\cite{halcon} application-ready  one-shot approach for object detection and pose estimation. It works by approximating local geometry with oriented point pairs and exhaustive feature matching between the scene and the object model.  PPF-based methods led the BOP challenge~\cite{bopchallenge} until very recently, when they were outperformed by deep learning methods. The disadvantages of the PPF methods is their dependence on depth information with per-point normals and a slow run time, which limits the potential applications.

Several attempts have been made to make deep learning-based pose estimation methods generalizable to new objects. NOCS~\cite{wang2019normalized}, followed by ~\cite{manhardt2020cps,chen2020category,li2022polarmesh}, proposed training a network to predict the poses of objects in a specific narrow class. Despite  advancements, the performance is still entirely dependent on the degree of similarity of objects within the object category. On the other hand, we make no such assumption because our approach explicitly employs the 3D object model during inference. Another line of research attempts to capitalize on similarities between objects in training and testing sequences. CorNet~\cite{pitteri2019cornet} used only corners to approximate object geometry.  Dense local shape descriptors for each object pixel were predicted in  \cite{pitteri20203d}, which could be matched with the object model. These methods needed a high degree of similarity between training and testing objects to work well. Sundermeyer et al.~\cite{sundermeyer2020multi} demonstrate that rotation estimation by template matching can generalize to new objects that were not seen during training, even if the feature extractor was trained on objects from a different dataset. The method, however, trains a 2D object detector on the target object classes. In contrast to them, our proposed method is fully one-shot and does not require any training on the target objects. The method of \cite{xiao2019pose} is more similar to our method. Here, the viewpoint is directly predicted by concatenating features extracted from the 3D model and the image. However, the method assumes that the object is already perfectly localized in 2D and does not estimate the full 6 DoF pose.\\

{\textbf{One-Shot Methods.}} 2D object detectors~\cite{girshick2014rich,girshick2015fast,ren2016faster,lin2017feature,liuSSDSingleShot2016,redmonyouonlylook2015,tian2019fcos} show excellent results when trained on target objects but cannot detect novel objects by design. One-shot methods~\cite{karlinsky2019repmet,michaelis2018one,caelles2017one,NIPS2019_8540,kang2019few} focus on generalization and allow for object detection of a new object, represented by a single template, in an input image. A common design principle is to  extract features from a template and a target image using a Siamese network and then match them to localize the object. The most related paper to our approach is OS2D~\cite{osokin2020os2d}, which is built around the ideas of feature matching for image alignment~\cite{rocco2017convolutional,Rocco18,rocco2018end}. It proposed to correlate template and image features using the sliding window approach. The key difference between our method and one-shot object detection methods is that we aim to explicitly use the full 3D information of the object mode.

\section{Methodology}

One-shot methods for 2D object detection use a target RGB image and a query template of the object of interest as input during inference, neither of which was seen during training. The inputs in our method are a target RGB image and a 3D model of the object.  The 3D model is represented by a set of 2D query templates rendered from various virtual camera viewpoints placed on a sphere around the object. Our pipeline consists of four stages as summarized in Figure~\ref{fig:pipeline} with stages object segmentation (1), template matching (2), 2D-2D matching (3), and pose estimation (4). Stages 3 and 4 can potentially be executed several times to produce multiple pose hypotheses.

In the following, we use $I$ of size $H\times W$ to denote an image, which implicitly depends on the object model $\mathcal{M}$ and its pose $T \in SE(3)$. 
A feature extractor $F_{FE}^k(I) \in \mathbb{R}^{H^k \times W^k \times D^k}$ uses a pre-trained network to extract feature maps of depth dimension $D^k$ from the input image. In the paper, we use pre-computed feature maps from several depth levels of the network, which are indexed by $\{ k \in \mathbb{N} \mid 1\leq k\leq N \}$.
$\mathit{\bar{F}}_{FE}^k(I) \in \mathbb{R}^{D^k}$ stands for a feature extractor which extends $\mathit{F}_{FE}^k$ by spatial averaging along height and width for each depth dimension of the feature map to produce a single vector of length $D^k$.

\subsection{One-Shot Segmentation}

\begin{figure}[!t]
    \centering
    \includegraphics[width=.8\linewidth]{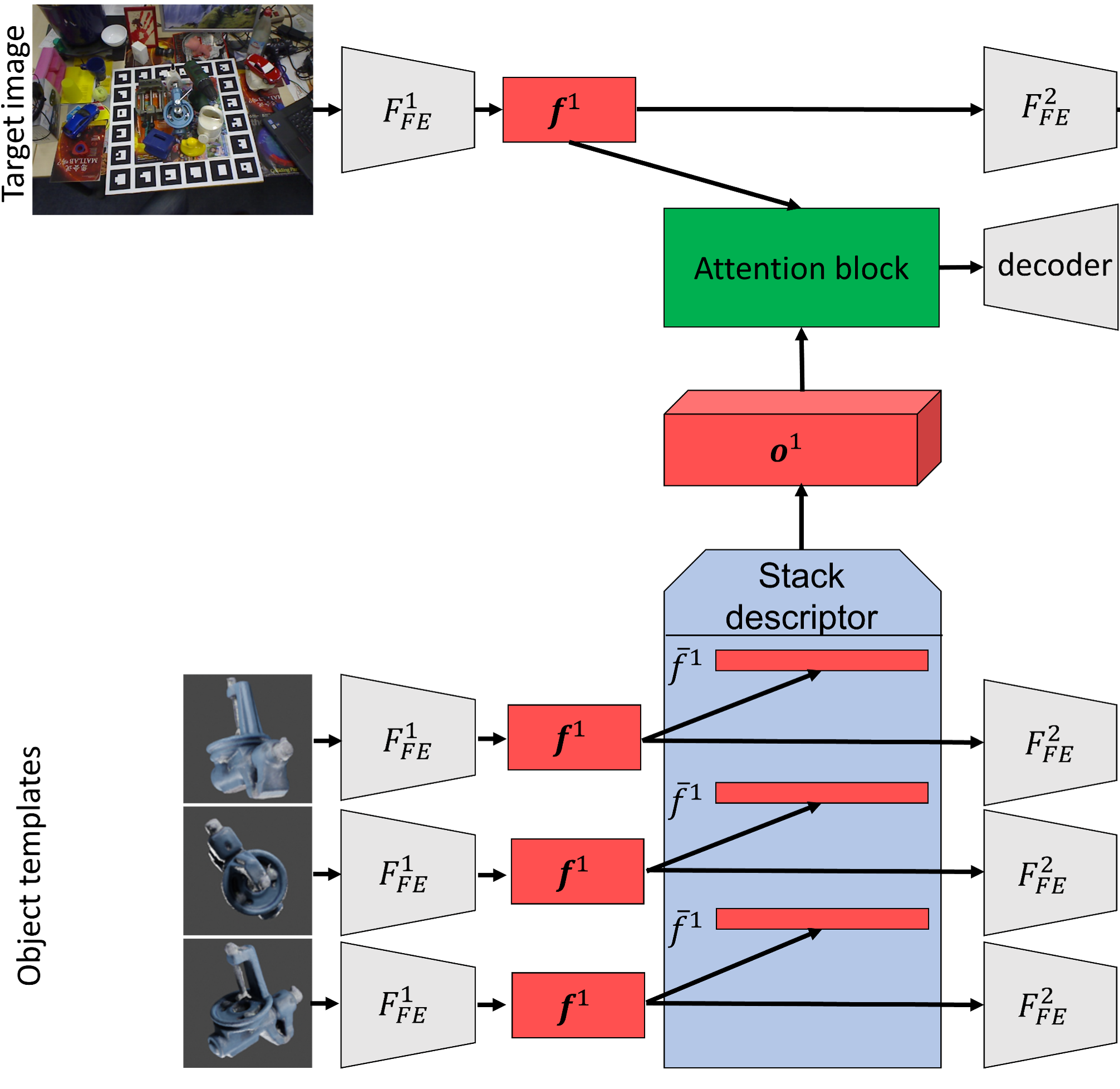}
    \vspace{-1em}
    \caption{Encoder of the stage 1 network. The network takes an input image and an object model, represented by a sparse set of templates, and outputs a binary segmentation of the target image. 
    The full detailed architecture is provided in the supplementary materials. }
    \label{fig:stage1_encoder}
    \vspace{-1.5em}
\end{figure}

The first stage network takes an image and a descriptor of the 3D model and predicts a binary segmentation mask indicating the location of the object' visible part. The core idea is to describe a textured CAD model using a set of viewpoint-based templates generated by rendering the object in various rotations. This brings the problem closer to the standard 2D one-shot methods. The key difference is that we compute a single descriptor based on pre-rendered templates, allowing the image to be matched to all  of the templates in a single shot, as opposed to the standard one-shot object detection, which treats each query template independently. Figure ~\ref{fig:stage1_encoder} depicts the overall architecture of the network.

\begin{figure}[!t]
    \centering
    \includegraphics[width=0.8\linewidth]{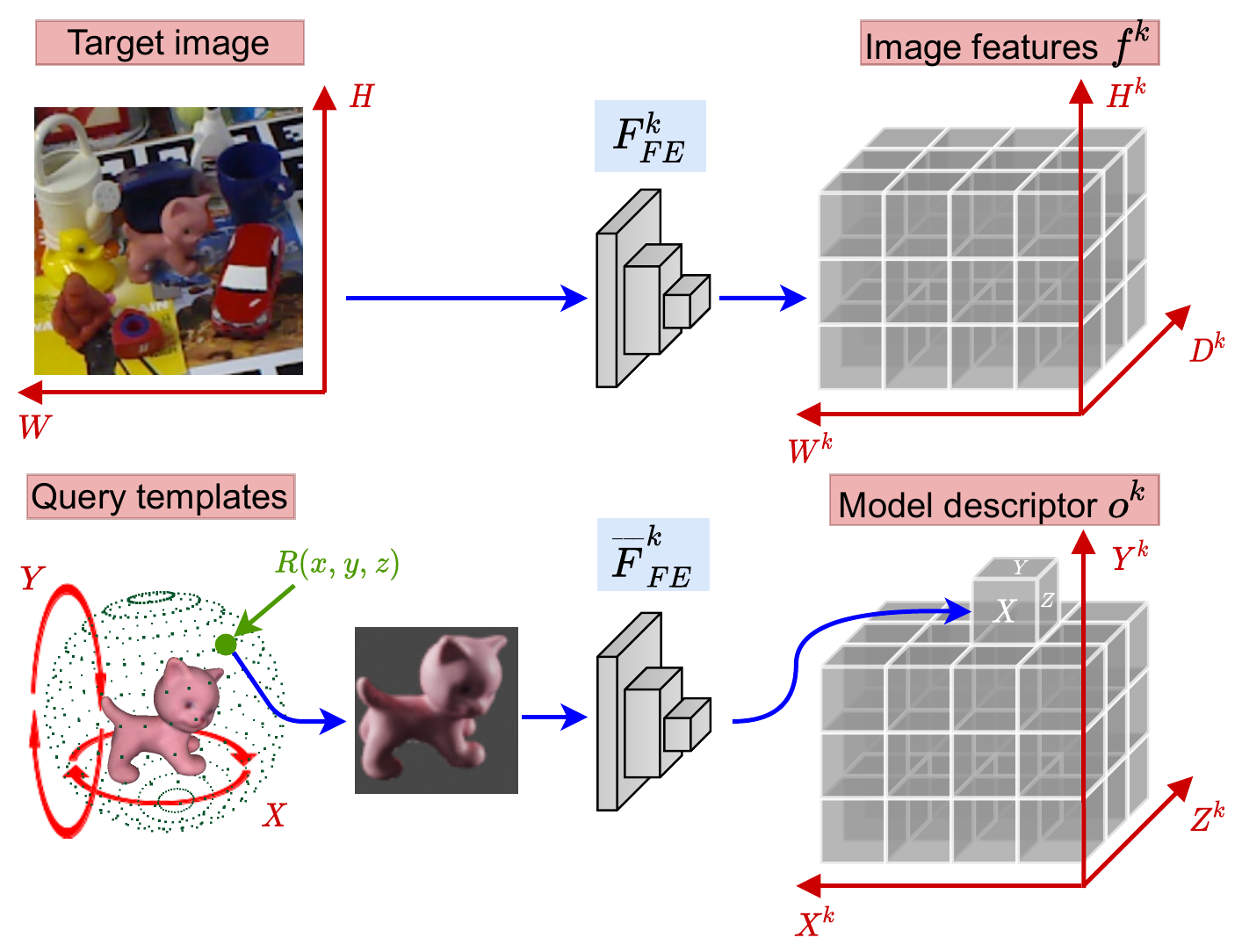}
    \vspace{-1em}
    \caption{Feature computation for a target image and a 3D model. The top row illustrates how an input target RGB image is converted to a 3D feature tensor $f^k$. The bottom row demonstrates how object templates sampled along azimuth (axis X), elevation (Y) and in-plane rotation axis Y are transformed to a corresponding dense 4D model descriptor $o^k$.}
    \label{fig:feature_computation_detection}
    \vspace{-1.5em}
\end{figure}

We build on the concept of dense feature matching first, which was first proposed in~\cite{rocco2017convolutional} and later extended for the task of object detection in~\cite{osokin2020os2d}. The basic idea is to compute per-pixel correlations between feature maps of the target image and the features from the object descriptor. Feature precomputation is visualized in Figure~\ref{fig:feature_computation_detection}. Pre-computed image features  $\mathbf{f}^{k} = \textit{F}_{FE}^k\left(I\right) \in \mathbb{R}^{H^k \times W^k \times D^k}$ and a 4D descriptor tensor $\mathbf{o}^{k} = \textit{F}_{FE}\left(\mathcal{M}\right)  \in \mathbb{R}^{X^k \times Y^k \times Z^k \times D^k}$ for the object $\mathcal{M}$ are compared. The 4D descriptor tensor $\mathbf{o}^{k}$ collects all templates rendered from the virtual viewpoints on the sphere around the object, where
the first two dimensions $(X, Y)$ stand for a camera position w.r.t. the object coordinate system using polar coordinates, while the third dimension $Z$ stands for in-plane rotations. The $4$th dimension $D^k$ refers to feature maps extracted from each template at multiple depth levels of the neural network indexed by $k$. Each element of the tensor is one viewpoint template represented with the corresponding feature vector. It is defined as $\mathbf{o}^{k}_{x, y, z} = \mathit{\bar{F}}_{FE}^k \left( I\left(R\left(x, y, z\right) \cdot \mathcal{M} \right) \right)$, where $R$ is a rotation matrix representing a virtual viewpoint on the sphere. Each pixel in the feature map $\mathbf{f}^{k}$ is matched to the entire object descriptor $\mathbf{o}^{k}$, resulting in a correlation tensor $\textbf{c}^k \in \mathbb{R}^{H^k \times W^k \times X^k \times Y^k \times Z^k}$. For a particular pixel $(h, w)$, the correlation  tensor value is defined as 
\begin{equation}
    \textbf{c}_{h, w, x, y, z}^k = \text{corr}\left(\mathbf{f}^{k}_{h, w},\mathbf{o}^{k}_{x, y, z}\right),
\end{equation}
where $\text{corr}$ denotes Pearson correlation. The correlation tensor is then flattened to a 3D tensor $\textbf{c}^k \in \mathbb{R}^{H^k \times W^k \times \left( X^k Y^k Z^k\right)}$. This way, each pixel of the target image feature map gets the list of all correlations of its feature vector with all the feature vectors of the descriptor. 

The flattened correlation tensor is used in two ways.  First, pre-computed correlations are used directly as the input to the decoder, as  in~\cite{osokin2020os2d,rocco2017convolutional,Rocco18,rocco2018end}. For that, the tensor $\textbf{c}^k$ is processed by a $1\times1$ convolutional layer to reduce the number of dimensions from $\left(X^k Y^k Z^k\right)$ to $L^k$. 
The correlations are also used to compute pixel-wise attention. Pixel-wise attention allows us to effectively incorporate the original image features into the feature tensor and use them for more precise segmentation.

Raw pixel-wise attention at the feature map level $k$ is defined simply as a sum of all $\left(X^k Y^k Z^k\right)$ correlations for a given pixel as
\begin{equation}
    A^k_{h, w} = \max \left\{0,  \sum\limits_{j = 1}^{(X^k Y^k Z^k)} \mathbf{c}^k_{h, w, j} \right\}.
\end{equation}

Since simple attention can be very noisy in the early layers compared to later layers, we propose to condition per-pixel attention of each particular level $k$ on the attention from the last level $k_l$, which tend to be more precise but have low resolution, as $\hat{A}^k_{h, w} = A^k_{h, w} \bigtriangledown\left(A^{k_l}\right)_{h^\prime, w^\prime}$. $\bigtriangledown$ denotes a bilinear upsampling of the attention  $A^{k_l}$ to the size of $A^k$. The values are then filtered by zeroing out attention values below the average value, resulting in cleaner and more precise attention maps, as illustrated in the supplementary material:
\begin{equation}
    \hat{A}^k_{h, w} = \begin{cases}
      \hat{A}^k_{h, w}, & \text{if $\hat{A}^k_{h, w} > \text{avg}_{h^\prime, w^\prime} \hat{A}^k_{h^\prime, w^\prime}$}\\
      0, & \text{otherwise}
    \end{cases} 
\end{equation}
 $A^{k_l}$ itself is thresholded but not conditioned on anything. All of the values are scaled to fall between 0 and 1. Image features are transformed using the attention maps as follows:
\begin{equation}
    \mathbf{\hat{f}}^k_{h, w} = \hat{A}^k_{h, w} \cdot \mathbf{f}^k_{h, w} - (1 - \hat{A}^k_{h, w}) \cdot \mathbf{f}^k_{h, w}.
\end{equation}

The attended features are then processed by a $1 \times 1$ convolutional layer to reduce dimensionality. Stacked $\mathbf{\hat{f}}^k$ and $\mathbf{c}^k$ are used jointly by the subsequent layers.  Overall, the decoder resembles the UNet~\cite{ronneberger2015u} approach of feature maps upsampling followed by convolutional layers until the initial image size is reached. The main distinction is that the network employs stacked $\mathbf{\hat{f}}^k$ and $\mathbf{c}^k$ at each level rather than skip connections. The network is trained to predict per-pixel probability that a pixel contains a visible part of the object. We used the Dice loss $\mathcal{L}_{Dice}$ \cite{milletari2016v} to handle imbalanced class data.

\subsection{Template Matching}

For the initial viewpoint estimation, we rely on template matching via deep manifold learning ,  which has been shown to scale well to a large number of objects~\cite{zakharov20173d,bui2018regression} and to generalize to new objects~\cite{sundermeyer2020multi} not seen during training. We rely on the same feature extraction network $\mathit{F}_{FE}$ but use only the features from the last layer. We also add one $1\times1$ convolutional layer to decrease the dimensions from $H \times W \times D$ to $H \times W \times D^\prime$.  Template features $\textbf{t} \in \mathbb{R}^{H\times W \times D^\prime}$ and image features $\textbf{f} \in \mathbb{R}^{H\times W \times D^\prime}$  are pre-computed from the foregrounds of the query templates and from the foreground of the detected object in the target image denoted with  using the segmentation predicted in the previous step respectively. Analogously to the first stage, similarity of two patches is estimated by computing per-pixel correlations between $\textbf{f}$ and $\textbf{t}$ using
\begin{equation}
    \text{sim}\left(\textbf{f}, \textbf{t}\right) = \sum_{h, w} \text{corr}\left(\textbf{f}_{h, w}, \textbf{t}_{h, w} \right).
\end{equation}
We train the network to increase similarity for patches which depict objects with very close rotations and at the same time penalize similarity for distant rotations. A modified triplet loss with dynamic margin is leveraged by optimizing
\begin{equation}
    \mathcal{L}_{triplets} = \max \left\{ 0, 1 - \frac{\text{sim}(\textbf{f}^{anchor}, \textbf{f}^{+})}{ \text{sim}(\textbf{f}^{anchor}, \textbf{f}^{-}) + \textit{m} } \right\},
\end{equation}
where $m$ is set to the angle between rotations of the object in the anchor and the puller patches. Using the terminology from~\cite{zakharov20173d},  $\textbf{f}^{anchor}$ is a descriptor of a randomly chosen object patch. $\textbf{f}^{+}$ corresponds to a puller - a template in the pose very similar to the pose in the anchor, while $\textbf{f}^{-}$ corresponds to the pusher with a dissimilar pose.   A query template with the highest similarity to the detected object in the target image  is chosen as a match at test time.

\begin{figure}[!t]
    \centering
    \includegraphics[width=.8\linewidth]{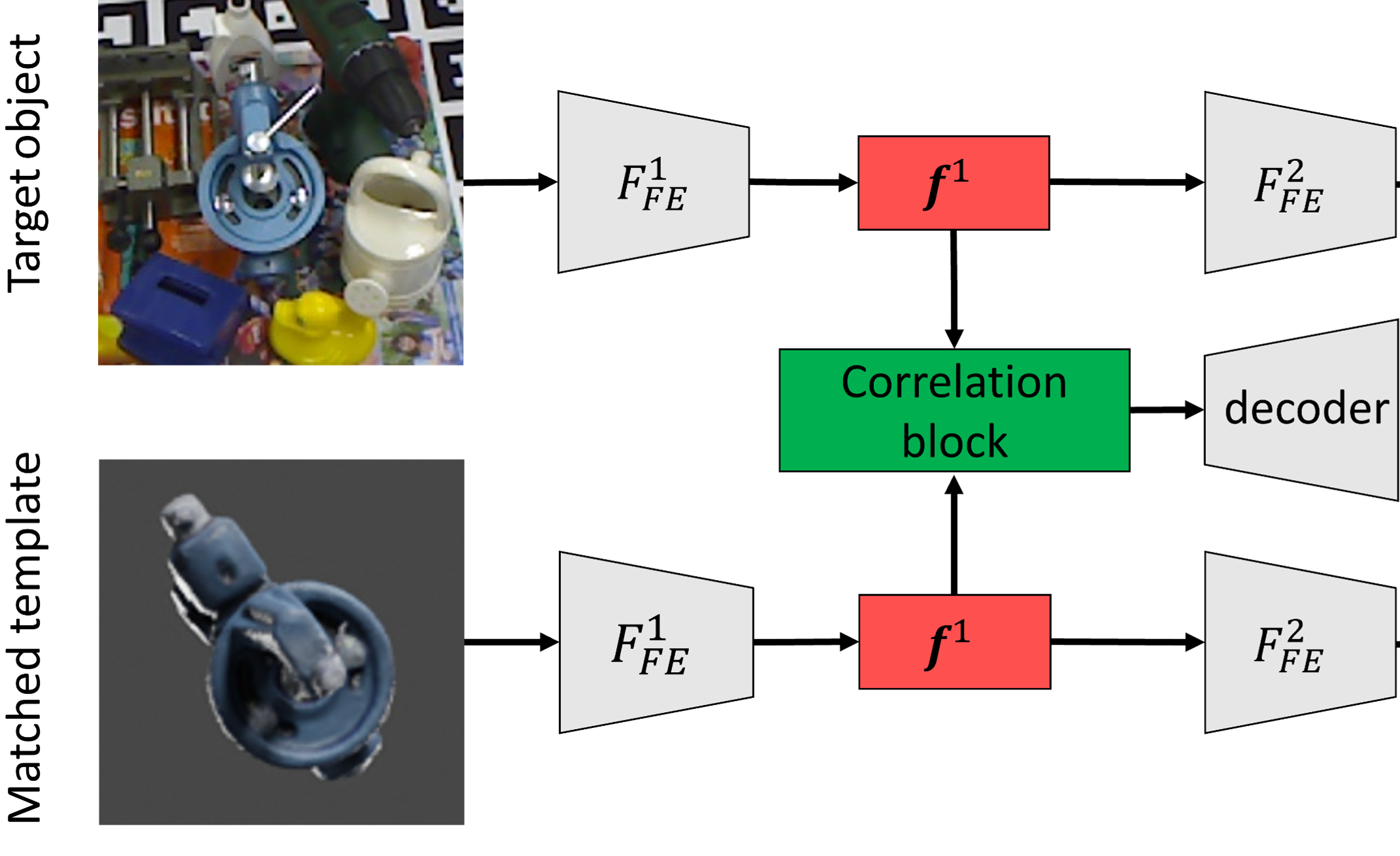}
    \vspace{-1em}
    \caption{Encoder of the stage 3 network. The network takes an input image with the detected object and a matched template. Its output is a pixel-wise binary segmentation and dense 2D-2D correspondences. 
    A detailed architecture is provided in the supplementary materials. 
    \label{fig:stage3_encoder}}
    \vspace{-1em}
\end{figure}

\subsection{One-Shot Dense Correspondence Estimation}

The goal of this stage is to establish 2D-3D correspondences between the image pixels and the object model. After the previous step, we have a patch with the detected object in unknown pose and a matched template in known pose. Establishing dense 2D-2D correspondences between the object patch and the template explicitly provides 2D-3D matches between the object pixels and the 3D object model. The correspondences can then be used to estimate the pose with PnP+RANSAC or Kabsch+RANSAC.

Similarly to the previous stages, the architecture of the 2D-2D matching follows the general idea of dense feature matching.
Each pixel of the feature map $\mathbf{f}^{k}$, representing the input image patch of the detected object, is matched with all pixels of the template feature map $\mathbf{t}^{k}$ to form the correlation tensor $\textbf{c}^k$.
The network then predicts three values for each pixel: a binary foreground/background segmentation mask and a coordinate of the corresponding pixel on the template. Figure~\ref{fig:stage3_encoder} depicts the architecture.
 
During training, a random object crop $I_{obj}$ with its associated pose $T_{obj} \in SE(3)$ is sampled from a synthetic dataset. Then, a random template $I_{tmp}$ is picked together with its pose $T_{tmp}  \in SE(3)$, so that $T_{obj}$ and $T_{tmp}$ are relatively close. Availability of object poses allows us to compute per-pixel 2D-3D correspondence maps in both patches. Let $C: \mathcal{M} \times SE(3) \rightarrow [0, 1]^{W \times H \times 3}$ denote 2D-3D correspondences for the object rendered in the given pose. 
Its inverse $C^{-1}$ recomputes correspondences with respect to the unnormalized object coordinates, corresponding to the actual 3D object coordinates. It allows us to define a 2D correspondence pair distance in the model's 3D coordinate space: 
\begin{multline}
    \textit{d}(p, p^\prime) = \norm{\textit{C}^{-1}\left(I_{obj} \right)_{p}  -  \textit{C}^{-1}\left( I_{tmp} \right)_{p^\prime}}_2
\end{multline}
where $p$ and $p^\prime$ are pixel coordinates in the image and template patches respectively. Ground truth dense 2D-2D correspondences are established by matching pixel pairs corresponding to the closest points in the 3D coordinate system of the model. For a point $p \in \mathcal{I}_{obj}$ its corresponding template point is computed as $\argmin_{p^\prime \in \mathcal{I}_{tmp}} \textit{d}\left(p, p^\prime\right)$.
We employ an outlier-aware rejection for 2D-2D correspondences with large 3D spatial discrepancy.
 
The segmentation loss is defined as a per-pixel Dice loss ($\mathcal{L}_{Dice}$). In addition, the network predicts a discrete 2D coordinate using a standard per-pixel cross-entropy classification loss denoted as $\mathcal{L}_{2D2D}$. 

\subsection{Pose Hypothesis Verification}
We propose an optional step for generating and verifying pose hypotheses. Its goal is to reduce the imprecision caused by incorrect initial viewpoint estimation. Pose hypotheses are generated by independently estimating 2D-2D correspondences from each of the top N matched templates and estimating poses from them. We greedily remove matched templates that are too close to each other to reduce run times and ensure more diverse poses. In practice, we filtered matches using a 15-degree threshold and picked top 25 templates from them. If depth is available, hypotheses are ranked based on the quality of fit of observed and rendered depth. In the RGB case, per-pixel correspondence error between the predicted correspondences and the rendered object is measured.
\section{Experiments}

We evaluate the proposed method on Linemod~\cite{hinterstoisser2012model} (LM), Occlusion~\cite{brachmann2014learning} (LMO), Homebrewed~\cite{Kaskman_2019_ICCV_Workshops} (HBD), YCB-V~\cite{xiang2018posecnn}. Evaluation on the TLESS dataset~\cite{hodan2017t} is provided in the supplementary material.
Each detector stage is trained separately for each target dataset, so that train and test objects differ. For example, we use train the networks used for experiments on the Linemod dataset using all objects from the Homebrewed and YCB-V datasets. Linemod's and Homebrewed's common objects are skipped during training. We used the synthetic PBR images provided by the organizers of the BOP challenge~\cite{bopchallenge} in all experiments. We denote methods, that used only PBR~\cite{bopchallenge} synthetic images, with "PBR", methods with custom synthetic images as "synt" and methods, which used a mix of real and synthetic data, as "mix". The standard ADD score~\cite{drost2010model} with the 10\% diameter threshold is reported for the Linemod dataset. The BOP Average Recall (AR) score~\cite{bopchallenge} is reported for the other datasets.

\begin{table}[!t]
	\centering
	\caption{2D detection results in comparison to YOLO~\cite{redmon2018yolov3}, trained on target objects, and one shot OS2D~\cite{osokin2020os2d} detectors on the BOP split of the test data~\cite{bopchallenge}. Results on the Homebrewed dataset~\cite{Kaskman_2019_ICCV_Workshops} are reported on the publicly available validation split. \label{tab:detection_comparison}}
	\resizebox{0.85\linewidth}{!}{%
		\begin{tabular}{c|c|ccc|c}
			\textbf{Dataset} & \textbf{Method} & \textbf{Precision} & \textbf{Recall} & \textbf{F1} & \textbf{Best Recall } \\
			\midrule
			\multirow{3}[2]{*}{LM} & YOLO  & 0.99  & 0.97  & 0.98  & 0.99 \\
			& Ours  & 0.47  & 0.86  & 0.61  & 0.86 \\
			& OS2D  & 0.28  & 0.2   & 0.23  & 0.57 \\
			\midrule
			\multirow{3}[2]{*}{LMO} & YOLO  & 0.69  & 0.67  & 0.68  & 0.85 \\
			& Ours  & 0.31  & 0.61  & 0.41  & 0.61 \\
			& OS2D  & 0.16  & 0.31  & 0.21  & 0.53 \\
			\midrule
			\multirow{3}[2]{*}{HBD} & YOLO  & 0.76  & 0.74  & 0.75  & 0.91 \\
			& Ours  & 0.43  & 0.73  & 0.54  & 0.73 \\
			& OS2D  & 0.24  & 0.29  & 0.26  & 0.44 \\
			\midrule
			\multirow{3}[1]{*}{YCB} & YOLO  & 0.72  & 0.84  & 0.78  & 0.98 \\
			& Ours  & 0.41  & 0.8   & 0.54  & 0.8 \\
			& OS2D  & 0.12  & 0.18  & 0.14  & 0.26 \\
		\end{tabular}%
	}%
\end{table}%

\subsection{2D Object Localization}

We compare our approach's 2D detection capabilities to two baselines. First, we compare it to OS2D~\cite{osokin2020os2d}, a state of the art one-shot object detection method. For fairness, we ran OS2D with exactly the same templates as our method. Second, we compare to YOLOv3~\cite{redmon2018yolov3}, which was trained separately for each scene using the synthetic PBR image~\cite{bopchallenge}. It establishes the upper bound for the object detection performance and demonstrates the recall achievable by a fully supervised 2D object detector trained on target objects. we cannot compute Mean Average Precision (mAP), because our localization network follows the semantic segmentation rather than the object detection paradigm. As an alternative, we chose a confidence threshold for each method on each dataset separately, maximizing the F1 score. We then report precision and recall that correspond to the optimal threshold as well as the highest possible recall achieved by the detector. Table~\ref{tab:detection_comparison} summarizes the findings. As expected, YOLO performs better than the proposed method and OS2D, both in terms of  precision and recall. Our method - not trained on the test objects - is outperformed by YOLO by 10\%-20\% in terms of best recall. At the same time, its recall is very close to the recall of YOLO with the confidence threshold corresponding to the best F1 score. Performance of OS2D is considerably worse, especially on more challenging datasets with more occlusion, e.g. LMO, HBD and YCB-V. Additionally, our method is ca.~$800\times$ faster with a runtime of only 25 milliseconds per object compared to 20 seconds per object for OS2D. This shows the advantages of the proposed method compared to state of the art in 2D one-shot detection.

\begin{table}[!t]
	\centering
	\caption{Percentages of correctly estimated poses w.r.t. the ADD on the Linemod~\cite{hinterstoisser2012model} dataset for methods trained on synthetic data. All methods apart from ours, PPF and PfS require prior training on RGB target objects.
		\label{tab:lm}}
	\resizebox{0.85\linewidth}{!}{%
		\begin{tabular}{c|cc|c|c}
			\toprule
			\textbf{Modality} & \textbf{Method} & \textbf{Refinement} & \textbf{ADD} & \textbf{Time (ms)} \\
			\midrule
			\multirow{8}[2]{*}{RGB} & DPOD~\cite{zakharov2019dpod}  & DL~\cite{zakharov2019dpod}    & 54.2  & - \\
			& Ours  & Mult. Hyp. & 43.6  & 1343 \\
			& DPOD~\cite{zakharov2019dpod}  & -     & 40.5  & 36 \\
			& OURS  & -     & 39.3  & 96 \\
			& SSD6D~\cite{kehl2017ssd} &  DL~\cite{manhardt2018deep}   & 34.1  & - \\
			& AAE~\cite{sundermeyer2018implicit}   & -     & 31.4  & 24 \\
			& PfS~\cite{xiao2019pose}   & -     & 22.5  & - \\
			& SSD6D~\cite{kehl2017ssd} & -     & 9.1   & - \\
			\midrule
			\multirow{7}[1]{*}{RGBD} & SSD6D~\cite{kehl2017ssd} &  ICP  & 90.9  & 100 \\
			& Ours  & Mult. Hyp. + ICP & 81.9  & 749 \\
			& Ours  & Mult. Hyp. & 80.1  & 722 \\
			& PPF~\cite{drost2010model}   & ICP   & 78.8  & - \\
			& Ours  & ICP   & 76.8  & 68 \\
			& Ours  & -     & 73.3  & 60 \\
			& AAE~\cite{sundermeyer2018implicit}   &  ICP  & 71.5  & 224 \\
		\end{tabular}
		\vspace{-2em}
	}
\end{table}%

\subsection{6 DoF Pose Results}

In this section, we assess the accuracy of poses estimated using our one-shot method. However, due to a lack of relevant work, it is not a straightforward task. Geometry-based deep learning methods~\cite{pitteri2019cornet,pitteri20203d} are evaluated on a single dataset using an old pose metric that the new state of the art methods do not report. Although Multi-Path AAE~\cite{sundermeyer2020multi} claims to be one-shot, it employs a 2D object detector trained on target objects. We, however, still compare to  Multi-Path AAE, as it  serves as a state of the art upper bound of what pose accuracy is achievable with deep learning methods capable of estimating the pose of novel objects not seen during training. when depth data is available, PPF results are reported, because it is another truly one-shot method that does not require training on target objects.  Other methods listed in the tables are explicitly trained on synthetic renderings of target objects, giving them an advantage over our method in terms of pose scores. Therefore, the results of our methods should not be directly compared to them; instead they should be used as a reference for what the standard 6 DoF pose estimation methods achieve. To summarize, we mainly compare our method to Multi-Path AAE on RGB images and to PPF on RGBD images.

\setlength{\tabcolsep}{5pt}
\begin{table}[!t]
	\centering
	\caption{Results on the Occlusion dataset~\cite{brachmann2014learning} reported according to the Average Recall (AR) metric of the BOP challenge~\cite{bopchallenge} on the BOP challenge subset of test images. All methods apart from ours and PPF~\cite{drost2010model} require prior training on target objects. }
	\label{tab:result_lmo}
	\resizebox{0.9\linewidth}{!}{%
		\begin{tabular}{c|cccc}
			\textbf{Method} & \textbf{Train data} & \textbf{Refinement} & \textbf{AR} & \textbf{Time (s)} \\
			\midrule
			CosyPose~\cite{labbe2020} & \multirow{6}[2]{*}{PBR} & -     & 0.633 & 0.550 \\
			CDPN~\cite{li2019cdpn}  &       & -     & 0.569 & 0.279 \\
			EPOS~\cite{hodan2020epos}  &       & -     & 0.547 & 0.468 \\
			Pix2Pose~\cite{hodan2020epos} &       & -     & 0.363 & 1.310 \\
			Pix2Pose~\cite{hodan2020epos} &       & -     & 	0.281 & 1.157 \\
			SSD6D~\cite{kehl2017ssd} &       & -     & 0.139 & - \\
			\midrule
			EPOS~\cite{hodan2020epos}  & \multirow{2}[2]{*}{synt} & -     & 0.443 & 0.487 \\
			DPOD~\cite{zakharov2019dpod}  &       & -     & 0.169 & 0.172 \\
			\midrule
			AAE ~\cite{sundermeyer2018implicit}  & \multirow{3}[2]{*}{mix} & ICP   & 0.237 & 1.197 \\
			Multi-Path AAE~\cite{sundermeyer2020multi} &       & -     & 0.217 & 0.200 \\
			AAE~\cite{sundermeyer2018implicit}   &       & -     & 0.146 & 0.201 \\
			\midrule
			Drost, PPF~\cite{drost2010model} & \multirow{8}[1]{*}{-} & ICP   & 0.527 & 15.947 \\
			Drost, PPF~\cite{drost2010model} &       & ICP, 3D edges & 0.492 & 3.389 \\
			Ours + Kabsch &       & Mult. Hyp. + ICP & 0.482 & 5.440 \\
			Ours + Kabsch &       & Mult. Hyp. & 0.462 & 5.355 \\
			Ours + Kabsch &       & ICP   & 0.432 & 0.560 \\
			Ours + Kabsch &       & -     & 0.393 & 0.475 \\
			Ours + PnP &       & Mult. Hyp. & 0.312 & 12.180 \\
			Ours + PnP &       & -     & 0.274 & 0.766 \\
		\end{tabular}%
		\vspace{-2em}
	}%
\end{table}%

In Table~\ref{tab:lm}, the quality of pose estimation  is reported using the ADD score for the Linemod dataset. We compare our approach to other methods that use only synthetic training data because they represent what can be accomplished without access to the training data from the target domain. Our method predicts very good poses despite not having been trained on Linemod objects. It significantly outperforms SSD6D~\cite{kehl2017ssd}, and outperforms AAE~\cite{sundermeyer2018implicit} and SSD6D with deep learning-based refinement~\cite{manhardt2018deep}, while falling short of DPOD~\cite{zakharov2019dpod} by around 1\%. Our method considerably outperform Pose from Shape~\cite{xiao2019pose} (PfS), which is another one-shot pose estimation method, even though PfS uses ground truth 2D detections and ground truth translations while estimating only rotation. Pose hypothesis verification raises the results by relative 10\% from 39.3 to 43.6. If depth data is available, the pose can be estimated directly using 3D-3D correspondences and the Kabsch algorithm. This nearly doubles our method's ADD score, putting it above AAE with ICP refinement. Further pose hypothesis verification improves the results by around relative 10\% putting it just above PPF. Segmentation of a single object takes 25 milliseconds, initial rotation approximation 10 milliseconds, 2D-2D matching 11 milliseconds. PnP takes 50 milliseconds on average, which makes the RGB-only pose estimation pipeline perform at approximately 10 FPS. If the Kabsch algorithm is used, the detector achieves 16 FPS. Additional ICP refinement on top of Kabsch slows down the detector to 14 FPS, which is still faster than other methods with ICP refinement. Pose hypothesis generation and verification in RGB takes around 1300 milliseconds and 550 milliseconds in depth images.

\setlength{\tabcolsep}{5pt}
\begin{table}[!t]
	\centering
	\caption{Results on the Homebrewed dataset~\cite{Kaskman_2019_ICCV_Workshops} reported according to the Average Recall (AR) metric of the BOP challenge~\cite{bopchallenge} on the BOP challenge subset of test images. All methods apart from ours and PPF~\cite{drost2010model} require prior training on target objects.}
	\vspace{-0.5em}
	\label{tab:result_hbd}
	\resizebox{0.9\linewidth}{!}{%
		\begin{tabular}{c|cccc}
			\textbf{Method} & \textbf{Train data} & \textbf{Refinement} & \textbf{AR} & \textbf{Time (s)} \\
			\midrule
			CDPNv2~\cite{li2019cdpn} & \multirow{7}[2]{*}{PBR} & -     & 0.722 & 0.273 \\
			CosyPose~\cite{labbe2020} &       & ICP   & 0.712 & 5.326 \\
			CDPNv2~\cite{li2019cdpn} &       & ICP   & 0.712 & 0.713 \\
			Pix2Pose~\cite{park2019pix2pose} &       & ICP   & 0.695 & 3.248 \\
			CosyPose~\cite{labbe2020} &       & -     & 0.656 & 0.417 \\
			EPOS~\cite{hodan2020epos}  &       & -     & 0.58  & 0.657 \\
			Pix2Pose~\cite{park2019pix2pose} &       & -     & 0.446 & 0.982 \\
			\midrule
			AAE ~\cite{sundermeyer2018implicit}  & \multirow{5}[2]{*}{synt} & ICP   & 0.506 & 1.352 \\
			CDPN~\cite{li2019cdpn}  &       & -     & 0.47  & 0.311 \\
			AAE~\cite{sundermeyer2018implicit}   &       & -     & 0.346 & 0.19 \\
			Multi-Path AAE~\cite{sundermeyer2020multi} &       & -     & 0.293 & 0.191 \\
			DPOD~\cite{zakharov2019dpod}  &       & -     & 0.286 & 0.18 \\
			\midrule
			Drost, PPF~\cite{drost2010model} & \multirow{8}[1]{*}{-} & ICP   & 0.671 & 144.029 \\
			Ours + Kabsch &       & Mult. Hyp. + ICP & 0.605 & 4.508 \\
			Drost, PPF~\cite{drost2010model} &       & ICP   & 0.603 & 1.659 \\
			Ours + Kabsch &       & ICP   & 0.581 & 0.438 \\
			Ours + Kabsch &       & Mult. Hyp. & 0.579 & 4.384 \\
			Ours + Kabsch &       & -     & 0.56  & 0.314 \\
			Ours + PnP &       & Mult. Hyp. & 0.492 & 8.183 \\
			Ours + PnP &       & -     & 0.464 & 0.503 \\
		\end{tabular}%
		\vspace{-2em}
	}%
\end{table}%


The results of our proposed method on the Occlusion (Table ~\ref{tab:result_lmo}) and Homebrewed (Table ~\ref{tab:result_hbd}) datasets show that it performs similarly to some of the methods that are trained on target objects with full supervision and trained separately for each object and scene. Even without refinement, our method outperforms Multi-Path AAE on both datasets. If depth data is available, our method falls short behind PPF only by a narrow margin while being an faster than the best performing PPF variant. On the YCB (Table ~\ref{tab:result_ycb}) dataset, Multi-Path AAE outperforms the proposed method (Our+PnP), but it is important to note that methods trained on real or mixed data perform considerably better than methods trained solely on synthetic data on this dataset. On the other hand, our method outperforms PPF by a large margin. The results on all datasets clearly show the competitive quality of object detection and pose estimation in the proposed one-shot method and demonstrate that it generalizes well to objects, which were not seen during training. Moreover, its performance matches the performance of some of the previous state of the art methods even though they were trained on the target objects.


\subsection{Ablation Studies}

\setlength{\tabcolsep}{5pt}
\begin{table}[!t]
	\centering
	\caption{Results on the YCB-V dataset~\cite{xiang2018posecnn}  reported according to the Average Recall (AR) metric of the BOP challenge~\cite{bopchallenge} on the BOP challenge subset of test images. All methods apart from ours and PPF~\cite{drost2010model} require prior training on target objects. }
	\vspace{-0.5em}
	\label{tab:result_ycb}
	\resizebox{0.9\linewidth}{!}{%
		\begin{tabular}{c|rccc}
			\textbf{Method} & \multicolumn{1}{c}{\textbf{Train data}} & \textbf{Refinement} & \textbf{AR} & \textbf{Time (s)} \\
			\midrule
			CDPNv2~\cite{li2019cdpn} & \multicolumn{1}{c}{\multirow{4}[2]{*}{PBR}} & ICP   & 0.532 & 1.034 \\
			EPOS~\cite{hodan2020epos}  &       & -     & 0.499 & 0.764 \\
			CDPNv2~\cite{li2019cdpn} &       & -     & 0.39  & 0.448 \\
			CosyPose~\cite{labbe2020} &       & -     & 0.574 & 0.342 \\
			\midrule
			EPOS~\cite{hodan2020epos}  & \multicolumn{1}{c}{\multirow{3}[2]{*}{synt}} & -     & 0.696 & 0.572 \\
			CDPN~\cite{li2019cdpn}  &       & -     & 0.422 & 0.295 \\
			DPOD~\cite{zakharov2019dpod}  &       & -     & 0.222 & 0.341 \\
			\midrule
			CosyPose~\cite{labbe2020} & \multicolumn{1}{c}{\multirow{7}[2]{*}{mix}} & ICP   & 0.861 & 2.736 \\
			CosyPose~\cite{labbe2020} &       & -     & 0.821 & 0.241 \\
			Pix2Pose~\cite{park2019pix2pose} &       & ICP   & 0.78  & 2.59 \\
			CDPNv2~\cite{li2019cdpn} &       & -     & 0.532 & 0.143 \\
			AAE~\cite{sundermeyer2018implicit}   &       & ICP   & 0.505 & 1.581 \\
			AAE~\cite{sundermeyer2018implicit}   &       & -     & 0.377 & 0.179 \\
			Multi-Path AAE~\cite{sundermeyer2020multi} &       & -     & 0.289 & 0.181 \\
			\midrule
			Ours + Kabsch &       & Mult. Hyp. + ICP & 0.572 & 2.606 \\
			Ours + Kabsch &       & ICP   & 0.565 & 0.302 \\
			Ours + Kabsch & \multicolumn{1}{l}{\textbf{-}} & Mult. Hyp. & 0.542 & 2.571 \\
			Ours + Kabsch &       & -     & 0.529 & 0.267 \\
			Drost, PPF~\cite{drost2010model} &       & ICP, 3D edges & 0.344 & 6.27 \\
			Ours + PnP &       & Mult. Hyp. & 0.332 & 5.389 \\
			Drost, PPF~\cite{drost2010model}&       & ICP, 3D edges & 0.33  & 1.282 \\
			Ours + PnP &       & \textbf{-} & 0.296 & 0.41 \\
		\end{tabular}%
	}
	\vspace{-2em}
\end{table}%

\begin{figure}[h]
	\vspace{-1em}
	\centering
	\begin{subfigure}{.49\linewidth}
		\centering
		\includegraphics[width=0.99\linewidth]{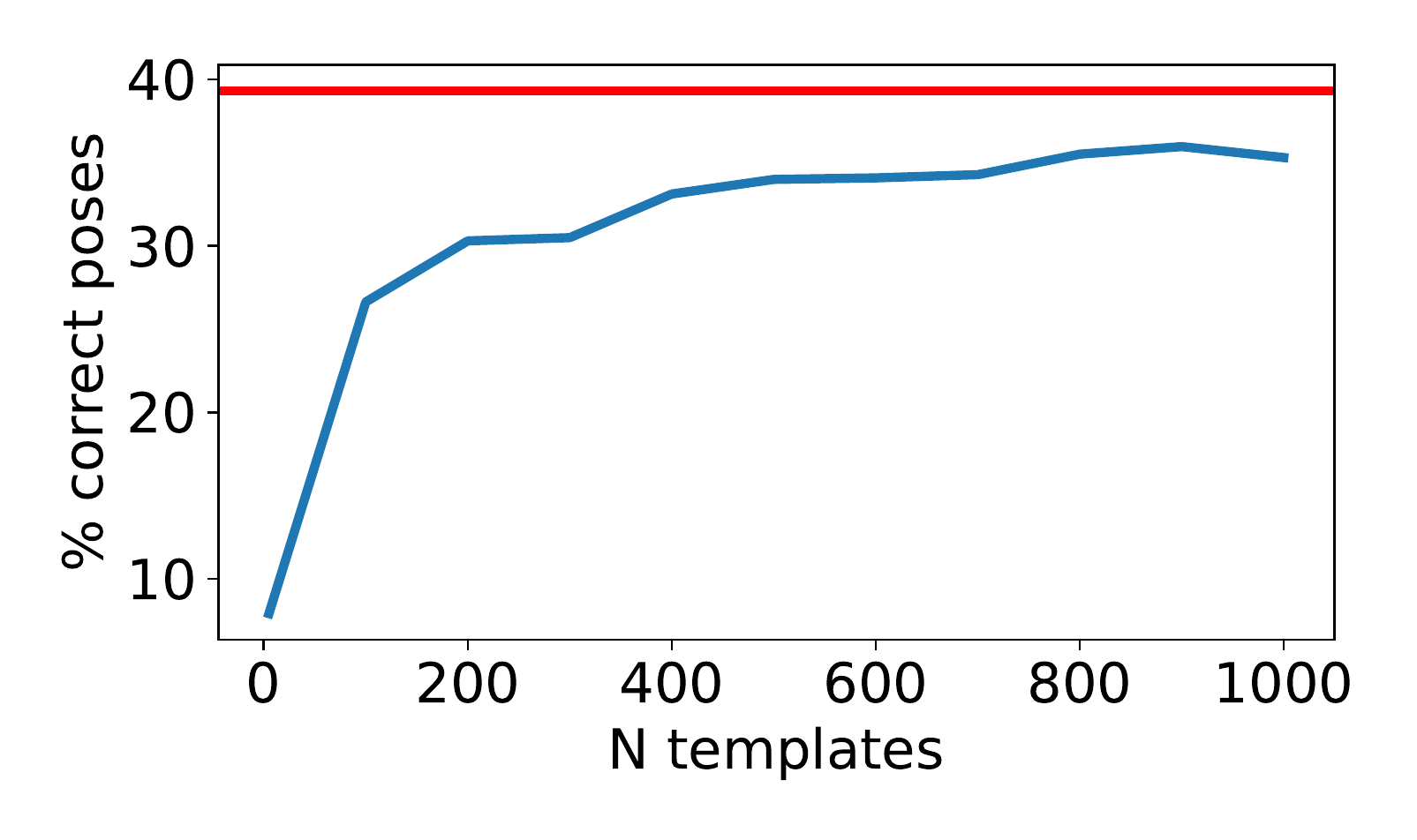}
		\caption{\label{fig:ablation_templates_n}}
	\end{subfigure}%
	\begin{subfigure}{.49\linewidth}
		\centering
		\includegraphics[width=0.99\linewidth]{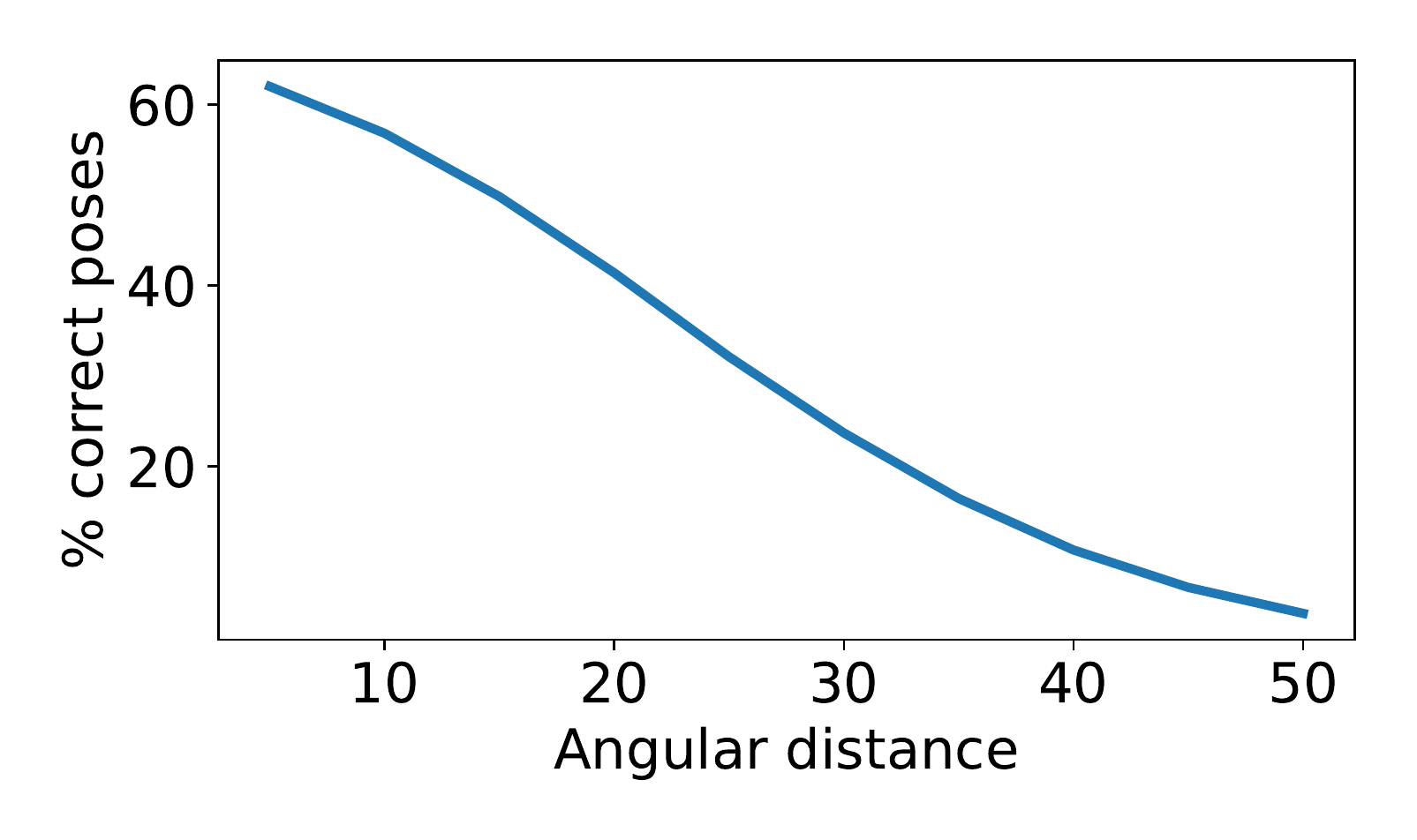}
		\caption{\label{fig:ablation_templates_n_angles}}
	\end{subfigure}%
	\vspace{-1em}
	\caption{Impact of the number of templates (a) and the angular distance between the ground truth and the matched template (b) on the final ADD score on Linemod dataset~\cite{hinterstoisser2012model}. \label{fig:ablation_templates}}
	\vspace{-2em}
\end{figure}

We conducted three main ablation studies to determine what factors contribute to the performance of our pipeline. Table~\ref{tab:ablation_2d} examines the architecture choices for the localization network, Table~\ref{tab:ablation_pose} analyses the pose estimation, while Figure~\ref{fig:ablation_templates} demonstrates the robustness of the method to a smaller number of templates and to larger angular errors between the ground truth and the matched templates.

Table~\ref{tab:ablation_2d} shows  that if we only use the feature correlation as in OS2D~\cite{osokin2020os2d,rocco2017convolutional}, the network performs the worst achieving only 54\%  recall and 51\% IoU. Only pixel-wise attention, on the other hand, yields comparable results. When both correlation and attention are used, network performance significantly increases which proves the effectiveness of the proposed architectural changes.

\setlength{\tabcolsep}{5pt}
\begin{table}[b]
	\footnotesize
	\centering
	\vspace{-2em}
	\caption{Object localization and segmentation for various configurations of the proposed localization network on Linemod~\cite{hinterstoisser2012model}.}
	\vspace{-1em}
	\label{tab:ablation_2d}
	\begin{tabular}{cc|ccc}
		\multicolumn{2}{c|}{\textbf{Configuration}} & \multirow{2}[2]{*}{\textbf{Precision}} & \multirow{2}[2]{*}{\textbf{Recall}} & \multirow{2}[2]{*}{\textbf{IoU}} \\
		\cmidrule{1-2}    Correlations & Attention &       &       &  \\ \midrule
		\checkmark     &       & 0.28  & 0.54  & 0.51 \\
		& \checkmark     & 0.34  & 0.61  & 0.55 \\
		\checkmark     & \checkmark     & 0.47  & 0.86  & 0.72 \\
	\end{tabular}%
\end{table}%

Table~\ref{tab:ablation_pose} snows an analysis of the impact of various stages on pose estimation.  
We start by replacing the first two stages of the pipeline with ground truth and then gradually replace it with actual predictions form our networks. ADD10 score is computed w.r.t. the number of correctly detected objects unless specified otherwise in "Recall", which multiplies the ADD score by the recall.   
The first line sets the upper bound for the pose estimation performance by effectively replacing the first two stages with ground truth and only using predictions from the 2D-2D matching network and PnP.
The second line introduces the template matching. The overall score decreases by only 2\% indicating that the second stage network performs precise template matching  given the GT segmentation masks and that the 2D-2D matching network is robust to larger angular differences between poses in the object patch and in the template. A large performance drop is observed in the third line, when GT segmentation is replaced with the predicted segmentation. This is further emphasized in the last row, where the ADD10 score is corrected by detector's recall. These results indicate that the main error comes from erroneous initial viewpoint estimation. Furthermore, improving  the predicted segmentation can further improve the overall performance of the one-shot pipeline by improving 2D recall and template matching. The table also shows that pose hypothesis verification helps filter out some of the erroneous poses and push the ADD score closer to the theoretical maximum of the method.

\setlength{\tabcolsep}{5pt}
\begin{table}[!t]
	\centering
	\caption{ADD10 score on the Linemod~\cite{hinterstoisser2012model} for different pipeline components.}
	\vspace{-1em}
	\label{tab:ablation_pose}
	\resizebox{0.9\linewidth}{!}{%
		\begin{tabular}{cccccc|c}
			\multicolumn{6}{c|}{Configuration}            & \multirow{2}[3]{*}{ADD10} \\
			\cmidrule{1-6}    GT segm. & Pred. Segm. & Closest tmpl. & Matched tmpl. & Recall & \multicolumn{1}{l|}{Mult. Hyp.} &  \\
			\midrule
			+     &       & +     &       &       &       & 60.9 \\
			+     &       &       & +     &       &       & 58.9 \\
			& +     &       & +     &       &       & 45.7 \\
			& +     &       & +     & +     &       & 39.3 \\
			\midrule
			& +     &       & +     &       & +     & 51.0 \\
			& +     &       & +     & +     & +     & 43.6 \\
		\end{tabular}%
	}%
	\vspace{-2em}
\end{table}%

We followed the example of Multi-Path AAE and used 90K templates for all experiments. 
Figure \ref{fig:ablation_templates_n}  shows the ADD score on Linemod if 5 to 1K templates randomly sampled from the full set are used. The red line corresponds to the ADD score of 39.3 with all 90K templates. OSOP reaches ADD of 35.3 with only 1K templates and 38.6 with 5K templates. Figure \ref{fig:ablation_templates_n_angles} shows how the angular distance between the gt rotation and the matched template affects the ADD score. OSOP requires a smaller number of templates and can estimate poses even from more distant template matches because of dense correspondence estimation.

\section{Limitations}

The method is predicated on the assumption that a pre-trained feature extractor computes distinctive features from the synthetic rendering of the object and the input image, and that the features have higher correlations for  templates and images that represent the object in similar poses. This assumption is influenced by the domain gap between real and synthetic images. This problem could potentially be remedied by unsupervised domain adaptation techniques.

\section{Conclusion}
We proposed a novel object detection and 6
DoF pose estimation method that generalizes well to the objects unseen during training. To the best of our knowledge,  it is the first one shot object detection and pose estimation method that does not impose any specific requirements for the objects. Our novel neural network architecture for one shot object localization performs significantly better and faster than an alternative 2D one shot detector. Our evaluation on Linemod, Occlusion, Homebrewed, YCB and TLESS datasets for pose estimation demonstrates the effectiveness of our method, which achieves similar results to methods trained on synthetic data. The proposed pipeline allows for a considerable reduction of time needed to prepare training data and to train the model, as these steps are not needed for new unseen objects. 

{\small
\bibliographystyle{ieee_fullname}
\bibliography{egbib}
}

\end{document}